\pdfoutput=1

\documentclass[11pt]{article}

\usepackage[final]{acl}

\usepackage{times}
\usepackage{latexsym}

\usepackage[T1]{fontenc}
 
\usepackage[utf8]{inputenc}

\usepackage{microtype}

\usepackage{inconsolata}

\usepackage{graphicx}

\usepackage{amsmath}
\usepackage{multirow}
\usepackage{multicol}
\usepackage{booktabs}
\usepackage{float}
\usepackage{placeins}
\usepackage{xcolor} 
\usepackage{colortbl}
\definecolor{myred}{RGB}{237, 211, 210} 
\definecolor{mygreen}{RGB}{198, 232, 206} 
\definecolor{myblue}{RGB}{218,232,252}

\title{Select, Read, and Write: A Multi-Agent Framework of \\ Full-Text-based Related Work Generation
}

\author{
 \textbf{Xiaochuan Liu}\textsuperscript{\rm 1,2,3},
 \textbf{Ruihua Song}\textsuperscript{\rm 1,2,3}\thanks{Corresponding authors: Ruihua Song and Xiting Wang.},
 \textbf{Xiting Wang}\textsuperscript{\rm 1,2,3}\footnotemark[1],
 \textbf{Xu Chen}\textsuperscript{\rm 1,2,3}
\\
 \textsuperscript{1}Gaoling School of Artificial Intelligence, Renmin University of China, Beijing, China \\
 \textsuperscript{2}Engineering Research Center of Next-Generation Intelligent Search \\and Recommendation, Ministry of Education \\
 \textsuperscript{3}Beijing Key Laboratory of Research on Large Models and Intelligent Governance
\\
 \texttt{
    \{liuxiaochuan,rsong,xitingwang,xu.chen\}@ruc.edu.cn
 }
}

\begin{document}
\maketitle
\begin{abstract}
Automatic related work generation (RWG) can save people’s time and effort when writing a draft of related work section (RWS) for further revision. 
However, existing methods for RWG always suffer from shallow comprehension due to taking the limited portions of references papers as input and isolated explanation for each reference due to ineffective capturing the relationships among them. 
To address these issues, we focus on full-text-based RWG task and propose a novel multi-agent framework. 
Our framework consists of three agents: a \emph{selector} that decides which section of the papers is going to read next, a \emph{reader} that digests the selected section and updates a shared working memory, and a \emph{writer} that generates RWS based on the final curated memory. 
To better capture the relationships among references, we also propose two graph-aware strategies for selector, enabling to optimize the reading order with constrains of the graph structure. 
Extensive experiments demonstrate that our framework consistently improves performance across three base models and various input configurations. 
The graph-aware selectors outperform alternative selectors, achieving state-of-the-art results.
The code and data are available at~\url{https://github.com/1190200817/Full_Text_RWG}.
\end{abstract}

\section{Introduction}
With the exponential growth of academic publications~\cite{wang2024autosurvey}, automatic related work generation (RWG) becomes more and more attractive to research communities because it can save time and effort in preparing the first draft of the related work section (RWS)~\cite{csahinucc2024systematic,martin2024shallow}. 
Although the RWG task has a long history~\cite{hoang2010towards} and the advancement of LLMs significantly improves the general ability of text understanding and generation, writing a good RWS is not trivial. Even for experienced researchers, they have to spend a bunch of time to draft the RWS after intensive reading of all references. They need to deeply comprehend the similarities and differences between references, organize them in a reasonable taxonomy, and position the current work by pointing out its novelty. However, existing methods are far from being as excellent as experienced researchers in writing RWS. 
There are at least two main challenges: 1) misinterpretations or hallucinations due to using limited portions of references (\emph{C1}) and 2) isolated explanation for each reference due to ineffective exploitation of the relationships (\emph{C2}).

\emph{C1}. 
Due to the limitations of input window sizes in language models, previous methods for RWG always rely on limited portions of references, such as abstracts~\cite{abura2020automatic,ge2021baco,li2022corwa,mandal2024contextualizing}, introduction and conclusion~\cite{chen2019automatic,deng2021automatic}, related work~\cite{xing2020automatic,ge2021baco}, or retrieved text spans~\cite{li2023cited,li2024explaining}, rather than leveraging the full texts. The lack of rich full-text information often prevents models from fully capturing the content and relationships among references, leading to frequent misinterpretations and hallucinations~\cite{xu2024hallucination}. However, full-text-based RWG task faces the challenge of limited context window size.
It often requires the inclusion of numerous lengthy references. 
Although models with long context windows have emerged (such as GPT-4o, 128K-token), directly feeding all textual data into the model in a single pass is not optimal. These models face diminishing performance when approaching their maximum context window~\cite{liu2024lost}. 

\emph{C2}. 
A high-quality RWS in academic writing needs to provide precise and in-depth comparisons across reference papers, highlight the novelty of the paper being written, and avoid isolated explanation of each reference~\cite{li2024related}. These criteria underscore an essential aspect of RWG: capturing and explaining the relationships among references. However, this is a common struggle in previous models, where loose relationships among references and isolated explanations for each reference are frequent issues~\cite{li2024explaining}. While some works leverage graph structure~\cite{chen2021capturing,wang2022multi} to model inter-paper relationships, they integrate graph structures implicitly, failing to effectively address the aforementioned issues.


We overcome the two challenges by proposing a multi-agent framework (\emph{C1}) and a graph-aware selector within the framework (\emph{C2}). We design our framework as a system comprising three agents: a \emph{selector}, a \emph{reader}, and a \emph{writer}. 
The first two agents work collaboratively and iteratively process input content while maintaining a shared working memory. 
The \emph{selector} decides the reading order of papers’ sections, and the \emph{reader} digests the selected content and updates the memory. Then the \emph{writer} generates the RWS based on the final curated memory. 
To better capture the relationships among references, we introduce the graph structure within our framework. We build two kinds of relationship graphs: a co-occurrence graph and a citation graph. 
Based on these graphs, we propose the graph-aware selector, which is able to explicitly obtain the structure of the graph and utilizes the relationships among references. 
\begin{figure*}[t]
\centering
  \includegraphics[width=0.9\linewidth]{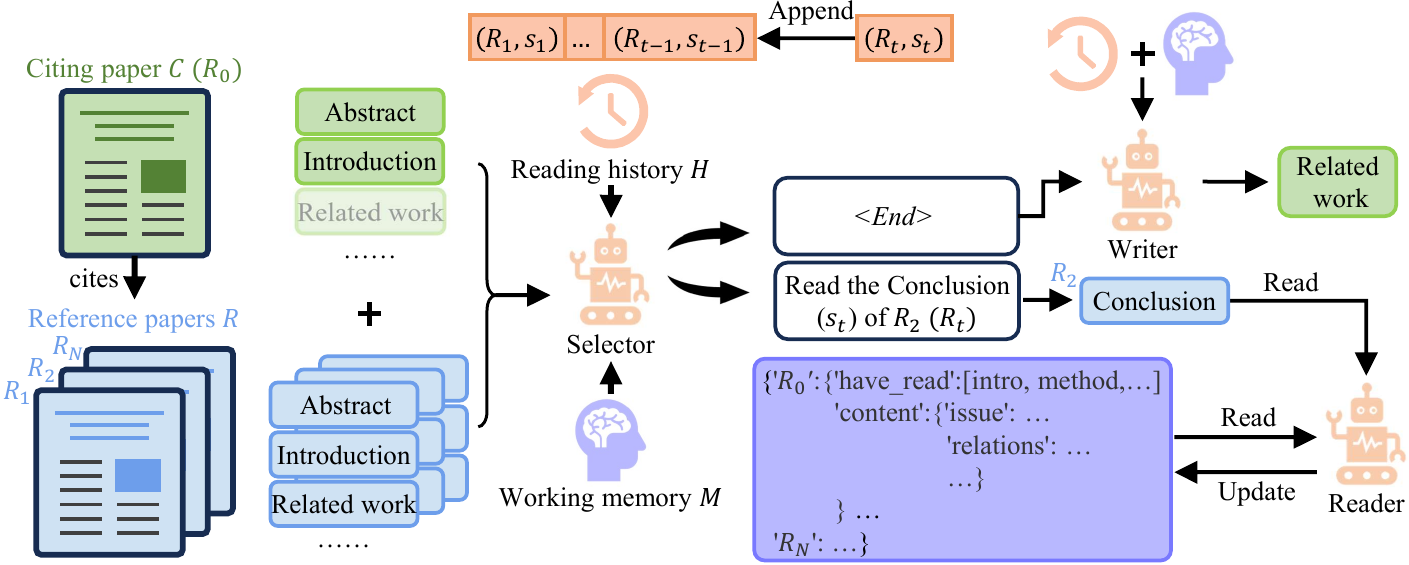}
  \caption{Overview of our multi-agent framework. The framework comprises a \emph{selector}, a \emph{reader}, and a \emph{writer}, which collaboratively read the papers and generate the related work section.}
  \label{fig:framework}
\end{figure*}
Extensive experiments demonstrate that our framework consistently improves performance across three base models (Llama3-8B, GPT-4o, and Claude-3-Haiku) in terms of LLM-based and graph-based metrics. 
Among selectors with different strategies, our graph-aware selectors perform the best. 

Our contributions can be summarized as follows:

\begin{itemize}
    \setlength{\itemsep}{0em}
    \item We propose a multi-agent framework for full-text-based RWG task. Our framework creatively delegates iterative reading to two distinct agents: the \emph{selector} and the \emph{reader}. And the \emph{writer} generates the final RWS.    
    \item We design two kinds of graphs and propose a graph-aware selector within our framework, which acts under the constraints of the graph.
    \item We conduct in-depth experiments on the impact of different selecting strategies and input configurations. Our framework consistently improves the performance across different configurations.
\end{itemize}
\section{Related Work}
\subsection{Related Work Generation}
Existing approaches for RWG can be categorized into two types: extractive and abstractive methods. Extractive methods focus on selecting key sentences from cited papers and concatenating them to form the related work section~\cite{hoang2010towards,hu2014automatic,wang2018neural,chen2019automatic,wang2019toc}. 
Recent RWG models predominantly adopt abstractive methods~\cite{chen2021capturing,chen2022target,liu2023causal}.
However, due to the limitations of input window sizes, these methods always rely on limited portions of reference papers,
such as abstracts~\cite{abura2020automatic,ge2021baco,li2022corwa,mandal2024contextualizing}, introductions and conclusions~\cite{chen2019automatic,deng2021automatic}, related work section~\cite{xing2020automatic,ge2021baco} or retrieved text spans~\cite{li2023cited,li2024explaining}. This lack of full-text information prevents models from fully capturing the content and relationships among references, leading to frequent misinterpretations and hallucinations~\cite{xu2024hallucination}. In addition, explaining the relationships among references is a critical aspect of RWG tasks. This is also a common struggle in previous models, where loose relationships among references and isolated explanations for each reference are frequent issues~\cite{li2024explaining}. Although some works attempt to model inter-paper relationships using relation graph~\cite{chen2021capturing} or knowledge graph~\cite{wang2022multi}, they integrate graph structures implicitly and the challenge remains largely unresolved. To address the above challenges, we focus on the full-text-based RWG task and incorporate explicit graph structure constraints within a multi-agent framework.

 
\subsection{Long-Sequence Modeling}
Extensive approaches are proposed to address the input length limitations of language models, which can be categorized into four types: context window scaling, recurrence-based methods, retrieval-based methods, and agent-based methods. 
Context window scaling methods extrapolate the positional embeddings~\cite{press2021train,chen2023extending} or employ modified self-attention mechanisms~\cite{beltagy2020longformer,guo2022longt5}. However, the attention mechanism may become less effective as sequence length increases~\cite{liu2024lost}. 
Recurrence-based methods use recursive mechanism to encode text, which are explored for different base models~\cite{miller2016key,chevalier2023adapting}. However, each recurrence step can introduce information loss. 
Retrieval-based methods retrieve relevant portions based on the query~\cite{izacard2021leveraging,wumemorizing}. However, such methods risk overlooking critical information. In agent-based frameworks, models operate as agents that dynamically read portions of the text and take flexible actions~\cite{nakano2021webgpt,yao2022webshop,chen2023walking,wang2024knowledge}. 
We adopt an agent-based framework, however, existing agent-based methods primarily focus on question answering (QA) tasks, where the agent only needs to locate an answer and a single agent suffices. In contrast, in RWG tasks, the reading order can impact the model’s understanding of the papers and their relationships. Our multi-agent framework creatively delegates the reading process to two distinct agents, enabling to optimize both reading order and updating memory.

\section{Problem Formulation}
Before introducing our framework, we first define the problem formulation and notations used throughout this paper.

The input to the task consists of two main components: the citing paper $C$, which represents the paper being written, and a set of reference papers $\mathcal{R} = \{ R_1, R_2, \dots, R_N \}$, where $R_i$ denotes a single reference paper and $N$ is the total number of references. For simplicity, $C$ is also denoted as $R_0$. Following prior RWG tasks, $\mathcal{R}$ is assumed to be given and corresponds to the references cited by the ground-truth related work section. Given the above inputs $\{R_0\} \cup \mathcal{R}$, the goal of the task is to generate a \textbf{related work section (RWS)} for $C$ that incorporates all references in $\mathcal{R}$ while maintaining coherence with the context of $C$. 

Since we focus on full-text-based RWG task, each reference paper $R_i$ contains its entire content, represented as $R_i = \{ s_{i,1}, s_{i,2}, \dots, s_{i,L_i} \}$, where $s_{i,j}$ denotes the $j$-th section of $R_i$, and $L_i$ is the total number of sections in $R_i$. 
Similarly, the input also includes all sections of the citing paper $R_0 = \{s_{0,1}, s_{0,2}, \dots, s_{0,L_0} \}$, except for the related work section, which is to be generated. 

\section{Multi-Agent Framework}

\begin{figure*}[t]
\centering
  \includegraphics[width=0.99\linewidth]{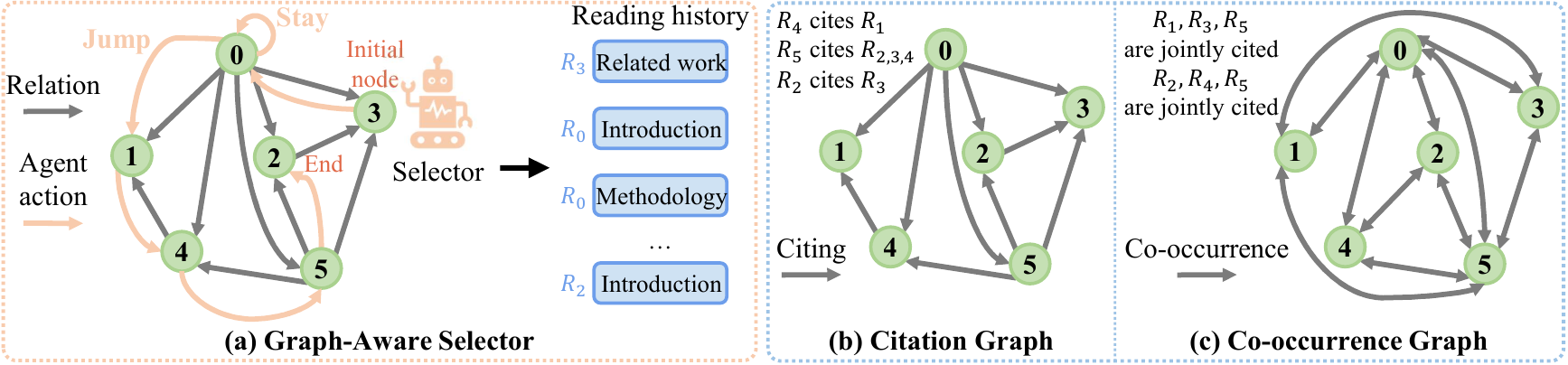}
  \caption{Illustration of our graph-aware selector. (a) Under the constraints of the graph structure, the selector selects either to continue reading the current paper or jump to an adjacent paper. We design two types of graphs: a (b) citation graph and a (c) co-occurrence graph.}
  \label{fig:deciders_impl}
\end{figure*}

\subsection{Overall Framework}
As shown in Figure~\ref{fig:framework}, our proposed framework is designed as a multi-agent system comprising three specialized agents: a \emph{selector}, a \emph{reader}, and a \emph{writer}. These agents work collaboratively to iteratively process input content while maintaining a shared working memory $M$ and a prior reading history $H$. The working memory $M$ is designed in a well-organized JSON format, storing key information deemed essential for drafting the RWS. The prior reading history $H$ records the sequence of previously read content in the form of tuples (\text{paper ID}, \text{section name}) in order to prevent cyclic reading of the input materials.

\textbf{Selector.}\quad
The selector is responsible for selecting the next section to read based on the abstracts of all papers $\{R_0\} \cup \mathcal{R}$, the current working memory $M_{t-1}$, and the reading history $H_{t-1}$. It outputs a tuple $(R_t, s_t)$, representing the selected paper ID and section name to be read at step $t$:
\begin{equation} 
(R_t, s_t) = \text{Selector}((s_{0,1}, \ldots, s_{N,1}), M_{t-1}, H_{t-1}),
\end{equation}
Here, $s_{0,1}$ to $s_{N,1}$ denote the abstracts of all papers.
Importantly, the selected section $(R_t, s_t)$ must not already exist in $H_{t-1}$.
When the selector determines that no further reading is necessary, it explicitly outputs a special termination symbol \emph{<End>} to conclude the iterative process.

\textbf{Reader.}\quad
The reader processes the content of the selected section $(R_t, s_t)$ and updates the working memory $ M_{t-1}$:
\begin{equation} M_t = \text{Reader}((R_t, s_t), M_{t-1}), 
\end{equation}
Given the task’s requirement to handle numerous lengthy references, the working memory $M$ can easily exceed the model’s context size limitation. To address this, we enforce an explicit size constraint on $M$ (e.g., 4096 tokens) and require the \emph{reader} to reorganize its contents at each step, discarding irrelevant information to maintain a concise and task-relevant memory.
After each iteration, the reading history $H$ is updated as follows:
$H_t = (H_{t-1}; (R_t, s_t))$. This iterative process continues until the selector signals termination.

\textbf{Writer.}\quad
Once the above iterative process concludes, the writer generates the final related work section based on the ultimate working memory $M_T$ and reading history $H_T$:
\begin{equation}
\text{RWS} = \text{Writer}(M_T, H_T),
\end{equation}
Here, $T$ represents the total number of iterations.

To guide the writer in understanding what constitutes a high-quality RWS in academic writing, we prompt the writer with explicit instructions (e.g., avoid isolated descriptions of each reference; explain the relationships between papers; group similar studies together). In addition, we provide the writer with an example of a well-crafted related work section to leverage the in-context learning capabilities of LLMs.
\subsection{Different Strategies for Selector}\label{deciders}

Since the reading order can impact the model’s understanding of the papers and their relationships, our framework is designed to allow diverse strategies for selector. In this paper, we investigate the following five distinct selectors, each offering a unique strategy for determining the reading order of papers and sections.

\textbf{Sequential Reading (SR).} 
The selector determines the reading order by following the papers' IDs sequentially. It reads each section of a paper in order before moving on to the next. Formally, the selector generates a reading history $H_T$ as:
\begin{equation}
\begin{split}
H_T = \{(R_0, s_{0,1}), &\ldots, (R_N, s_{N,L_N})\} ,
\end{split}
\end{equation}
Here, $s_{i,j}$ denotes the $j$-th section of paper $R_i$.

\textbf{Random Reading (RR).}
Sequential reading may introduce biases due to the fixed order of reading, such as prioritizing earlier-read papers. To mitigate the potential bias, we implement a random reading strategy. The selector shuffles the sequential reading history into a random order:
\begin{equation}
\begin{split}
     H_T = \text{shuffle}(\{(R_0, s_{0,1}), \ldots, (R_N, s_{N,L_N})\}) ,
\end{split}
\end{equation}

\textbf{Vanilla LLM-Based Selector (Vanilla).}
We explore a vanilla LLM-based selector that dynamically determines the reading order. At each step $t$, the selector selects the next paper and section as:
\begin{equation}
(R_t, s_t) = \text{LLM}((s_{0,1}, \ldots, s_{N,1}), M_{t-1}, H_{t-1}) ,
\end{equation}
This implementation takes advantage of the contextual reasoning abilities of LLMs to adaptively prioritize reading based on the task requirements.

\textbf{Graph-Aware Selector.}
Understanding the relationships among references is crucial for RWG tasks. The graph structure is an intuitive way to describe relationships. Therefore, we propose a novel graph-aware selector, which constrains the reading order within the graph, enabling the selector to capture the relationships among papers. Specifically, we propose building two types of graphs: a co-occurrence graph and a citation graph.

\noindent\textbf{Co-occurrence Graph (Graph-Co).}\quad In practice, the RWS of reference papers $\mathcal{R}$ can provide valuable guidance for writing the RWS of the citing paper $R_0$. If the RWS of a reference discusses certain papers together, it is likely that these papers share a strong connection. To model this, we construct a co-occurrence graph (as shown in Figure~\ref{fig:deciders_impl}(c)), where each node represents a reference paper, and an edge between two nodes indicates that the two papers are co-occurred in the same sentence of a prior paper’s RWS. For convenience, we define the co-occurrence graph as a directed graph $G_{co} = (V_{co}, E_{co})$, vertices $V_{co} = \{R_0\} \cup \mathcal{R}$, edges $E_{co} = \{(R_i, R_j) \mid R_i \text{ and } R_j \text{ are} \text{ jointly} \text{ cited} \text{ in } R_k \text{’s RWS}\}$. Importantly, the citing paper $R_0$ is assumed to be connected to all the reference papers in the graph to ensure its accessibility. The co-occurrence graph can effectively capture the implicit relationships among papers as exhibited in prior works.

\noindent\textbf{Citation Graph (Graph-Ci).}\quad For all papers $\{R_0\} \cup \mathcal{R}$, there is a citation graph $G_{ci} = (V_{ci}, E_{ci})$, vertices $V_{ci} = \{R_0\} \cup \mathcal{R}$, edges $E_{ci} = \{(R_i, R_j) \mid R_i \text{ cites } R_j\}$. Citation relationships between papers can provide a more direct and explicit way to model inter-paper connections. Importantly, the citing paper $R_0$ is also included in the graph and cites all the reference papers in $\mathcal{R}$. 

As shown in Figure~\ref{fig:deciders_impl}(a), the graph-aware selector begins by selecting an initial paper $R_{\text{init}}$ based on $G$. At each step $t-1$, the selector is positioned at a paper $R_{t-1}$ and operates within its one-hop subgraph $G_{t-1} = (V_{t-1}, E_{t-1})$, defined as:
\begin{equation}
\begin{split}
    V_{t-1} = \{R_{t-1}\} \cup \{R_i \mid & (R_i, R_{t-1}) \in G \text{ or } \\
                         &(R_{t-1}, R_i) \in G\},
\end{split}
\end{equation}
\begin{equation}
\begin{split}
    E_{t-1} = \{(R_i, R_j) \mid & R_i, R_j \in V_{t-1}, \\
                                &(R_i, R_j) \in G\} ,
\end{split}
\end{equation}
Within this subgraph, the selector selects either to continue reading the current paper or jump to an adjacent paper:
\begin{equation}
\begin{split}
    (R_t, s_t) = \text{Selector}(&(s_{0,1}, \ldots, s_{N,1}),\\
                                &M_{t-1}, H_{t-1}, G), \\
                                &R_t \in V_{t-1} ,
\end{split}
\end{equation}
We grant the selector access to the entire graph $G$ as well as the abstracts of all papers, enabling it to make globally informed decisions.

\section{Experiments}
\begin{table*}[t]
  \centering
  \setlength{\tabcolsep}{4.8pt} 

  \begin{tabular}{l|ccc|cccc}
    \toprule
    \multirow{3}{*}{\textbf{Model}} & \multicolumn{3}{c}{\textbf{Graph-based Metrics}} & \multicolumn{4}{c}{\textbf{LLM-based Evaluation}} \\
    & Avg. & Avg. Node & Clustering & \multirow{2}{*}{Coverage} & \multirow{2}{*}{Logic} & \multirow{2}{*}{Relevance} & \multirow{2}{*}{Overall} \\
    & Edges & Degree & Coefficient &&&& \\
    \midrule
    \multicolumn{8}{l}{\textit{Abstract-based RWG Models}} \\
    STK5SciSumm & 0.0 & 0.0 & 0.0 & 1.02 & 1.04 & 2.18 & 1.41 \\
    PRIMERA & - & - & - & 1.60 & 1.66 & 3.42 & 2.23 \\
    Llama3-8B &1.000&0.348&0.054&2.64&3.16&4.04&3.28 \\
    Claude-3-Haiku &1.729 & 0.448 & 0.084& 2.84 & 3.40 & 4.10 & 3.45 \\
    GPT-4o &1.180&0.439&0.057&3.16&3.70&4.22&3.69 \\
    \citeauthor{li2024explaining} & 1.810 & 0.629 & 0.113 & 3.25 & 3.74 & 4.30 & 3.76\\
    \midrule
    \multicolumn{8}{l}{\textit{Retrieval-based Full-Text RWG Models}} \\
    STK5SciSumm + GO & 0.0 & 0.0 & 0.0 & 1.08 & 1.14 & 2.48 & 1.57 \\
    PRIMERA + GO & - & - & - & 1.78 & 1.72 & 3.42 & 2.31 \\
    Llama3-8B + GO & 1.511 & 0.350 & 0.054 & 2.68 & 3.16 & 4.02 & 3.29 \\
    Claude-3-Haiku + GO & 2.308 & 0.530 & 0.100 & 2.90 & 3.48 & 4.18 & 3.52 \\
    GPT-4o + GO &1.611 & 0.535 & 0.096 & {3.22} & \underline{3.76} & {4.28} & {3.75}\\
    \midrule
    \multicolumn{8}{l}{\textit{LLMs with Extended Context Windows}} \\
    Claude-3-Haiku & {2.344} & \underline{0.869} & 0.097 & 2.34 & 3.32 & 3.74 & 3.13 \\
    GPT-4o & 1.244 & 0.624 & 0.136 & 3.18 & 3.66 & 4.20 & 3.68 \\
    \midrule
    \multicolumn{8}{l}{\textit{Ours}} \\
    $\text{Llama3-8B}_\text{ Graph-Co}$ & 1.162 & 0.644 & 0.135 & 2.74 & 3.20 & 3.98 & 3.31 \\
    $\text{Llama3-8B}_\text{ Graph-Ci}$ & 1.410 & 0.651 & {0.154} & 2.80 & 3.34 & 4.18 & $\text{3.44}^{*}$ \\
    $\text{Claude-3-Haiku}_\text{ Graph-Co}$ & \underline{2.840} & 0.832 & \underline{0.210} & 2.98 & 3.48 & {4.22} & 3.56 \\
    $\text{Claude-3-Haiku}_\text{ Graph-Ci}$ & \textbf{3.240} & \textbf{0.942} & \textbf{0.231} & 3.00 & 3.62 & 4.22 & $\text{3.61}^{*}$ \\
    $\text{GPT-4o}_\text{ Graph-Co}$ & 1.900 & 0.649 & 0.123 & \underline{3.28} & 3.74 & \underline{4.34} & \underline{3.79} \\
    $\text{GPT-4o}_\text{ Graph-Ci}$ &2.125&0.667&0.128&\textbf{3.32}
&\textbf{3.86}&\textbf{4.44}&$\textbf{3.87}^{*}$ \\
    \bottomrule
  \end{tabular}
  \caption{Performance of different models on the OARelatedWork dataset. The best and runner-up are in \textbf{bold} and \underline{underlined}. Graph-based metrics for the PRIMERA model are not reported because its generated results do not distinguish between different references, making it infeasible to construct the corresponding graph. “*” indicates statistically significantly better than the corresponding strongest baseline with Paired t-test $p < 0.05$.}
  \label{tab:main_results}
\end{table*}
\subsection{Experiment Setup}
\noindent\textbf{Dataset.}\quad 
We utilize OARelatedWork dataset~\cite{docekal2024oarelatedwork}, currently the only dataset supporting full-text-based RWG. 
It has an average input length of 70k tokens in the test set. 
Unlike other datasets that are often domain-specific and focus on computer science~\cite{lu2020multi,chen2022target}, OARelatedWork is an open-domain dataset.
Due to the substantial size of the dataset, all our experiments are conducted on 10\% of the dataset.

\noindent\textbf{Implementation Details.}\quad 
We experiment with three advanced LLMs in our multi-agent framework: one open-source model Llama3-8B and two closed-source models Claude-3-Haiku~\footnote{Version claude-3-haiku-20240307} and GPT-4o~\footnote{Version gpt-4o-2024-08-06}. We select Llama3-8B, Claude-3-Haiku, and GPT-4o as base models because they represent different tiers of current LLM capabilities (weak, medium, and strong). By demonstrating consistent performance improvements across models with varying capabilities, we can validate the generalizability of our framework. While the use of closed commercial LLMs is common in NLP research, it poses challenges for reproducibility. To address this concern, we ensure that all experiments conducted with Llama3-8B are performed on-site, providing strict reproducibility. As detailed in Section~\ref{deciders}, our framework implements five distinct variants of the selector.
These variants are distinguished using subscripts throughout the paper. 
The prompts and implementation details for each agent are provided in the Appendix~\ref{app:prompt}.

\noindent\textbf{Baselines.}\quad
We compare our framework against baselines of three categories: 1) \textbf{Abstract-based RWG Models.} Due to the input length limitations of language models, most previous works generate RWS solely based on the abstracts. We take several state-of-the-art models as our baselines, including the traditional language models PRIMERA~\cite{xiao2022primera} and STK5SciSumm~\cite{to2024skt5scisumm}, advanced LLMs (Llama3-8B, Claude-3-Haiku, and GPT-4o), and LLM-prompting method~\cite{li2024explaining}. 2) \textbf{Retrieval-based Full-Text RWG Models.} Many studies address the challenge of processing long inputs by leveraging retrieval-based methods~\cite{izacard2021leveraging,wumemorizing}. In RWG tasks, the Greedy Oracle (GO)~\cite{nallapati2017summarunner} is a popular choice for selecting sentences. 
The selected sentences are then used as input to fit the context window of the model. We take the same models mentioned in abstract-based RWG category but extend their input to include sentences selected by the GO. 3) \textbf{LLMs with Extended Context Windows.} Certain advanced LLMs are equipped with long input windows, enabling them to process the full-text of all references simultaneously. We choose Claude-3-Haiku (200K-token) and GPT-4o (128K-token) as baselines.

\subsection{Metrics}
\begin{figure*}[t]
\centering
  \includegraphics[width=0.92\linewidth]{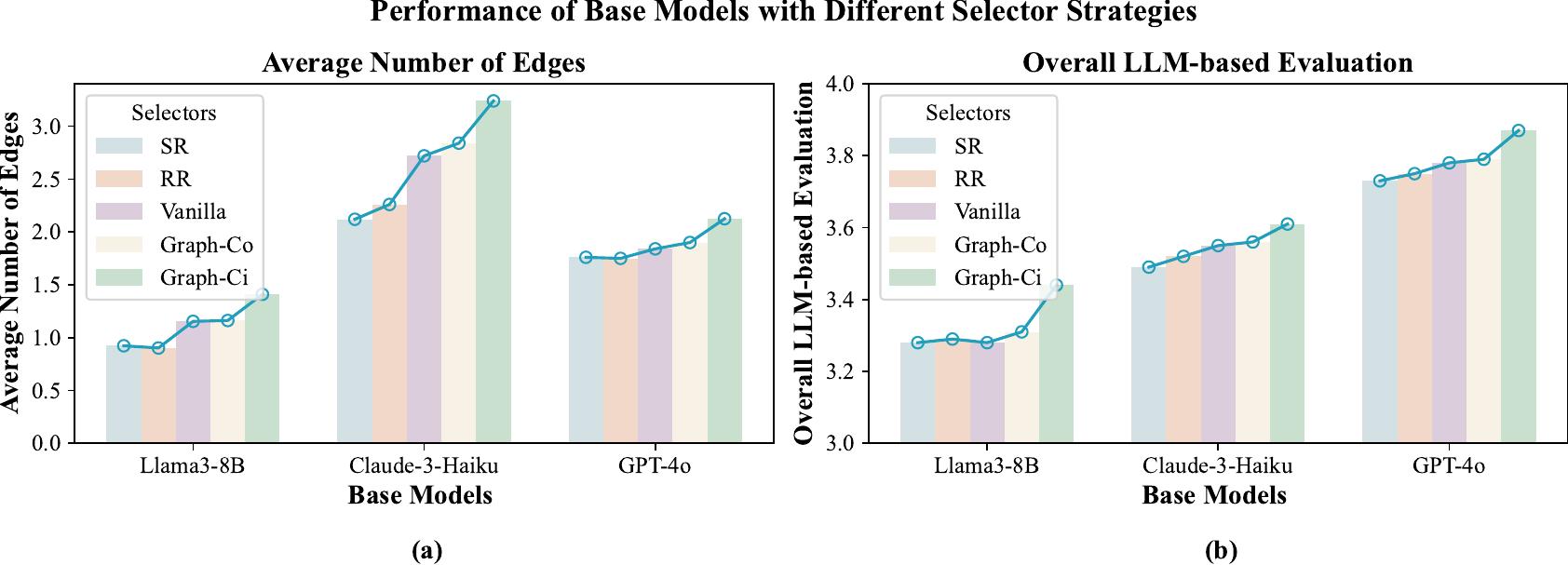}
  \caption{Performance comparison of five different selector strategies across three base models: (a) average number of edges in graph-based metrics, (b) overall LLM-based evaluation.}
  \label{fig:deciders}
\end{figure*}
To avoid the poor correlation with human judgments in traditional automatic metrics~\cite{chen2024rethinking}, we choose two kinds of evaluation methods specifically for the RWG task.

\textbf{Graph-based Metrics.} To evaluate how well the generated RWS integrates and relates references, we adopt graph-based metrics~\cite{martin2024shallow}. A co-occurrence graph is constructed from the generated RWS, where each node represents a reference paper, and edges indicate that two references are jointly cited in a single sentence. A denser graph can reflect a more interrelated explanation of the references. We select three simple yet insightful graph statistics as metrics: \textbf{Average Number of Edges}, \textbf{Average Node Degree}, and \textbf{Clustering Coefficient}. Clustering coefficient can evaluate the tendency of references to form tightly connected clusters and avoid overestimating quality for connections between unrelated references. 

\textbf{LLM-based Evaluation.} Previous LLM-based methods for evaluation predominantly focus on linguistic quality~\cite{ge2021baco, li2022corwa}.
To provide more accurate evaluations tailored to RWG tasks, we carefully design three metrics (inspired by~\citeauthor{wang2024autosurvey})—coverage, logic, and relevance—to assess the generated content’s alignment with the essential characteristics of a high-quality RWS. \textbf{Coverage}: whether the generated RWS covers all key topics and provide detailed and thorough discussions about the references. 
\textbf{Logic}: whether the RWS is tightly structured and logically coherent, with content arranged in a clear and reasonable manner. \textbf{Relevance}: whether the RWS aligns with all papers and avoids hallucinations or factual inaccuracies.
Each metric is scored on a 5-point scale. To enhance the accuracy and consistency, we employ chain-of-thought (CoT) prompting~\cite{yu2023towards} with explicit scoring criteria. 
To mitigate potential biases introduced by the preferences of individual LLMs, we utilize three advanced LLMs: GPT-4o~\footnote{A distinct version, gpt-4o-2024-05-13}, Claude-3.5-haiku, and Gemini-1.5-Pro. The final results are the average of these three models. Details on the prompt design and scoring criteria for each metric are provided in the Appendix~\ref{app:prompt}.

\subsection{Main Results}
We compare our framework with three types of baselines and present the results in Table~\ref{tab:main_results}. We report the performance of our framework with graph-aware selectors, as they achieve the best performance. The key findings from the table are as follows:
(1) \textbf{Full-text-based RWG models outperform abstract-based models.} The performance of full-text-based RWG models (including retrieval-based models and our framework) is significantly better than that of abstract-based RWG models. This trend holds consistently across five baselines and two kinds of evaluation metrics. It validates our motivation for full-text-based RWG tasks.
(2) \textbf{Feeding all textual data in a single pass is not optimal.} While many advanced LLMs claim to handle long inputs, their methods for extending input windows often come at the cost of performance. As shown in Table~\ref{tab:main_results}, for GPT-4o and Claude-3-Haiku, feeding all the content does not perform as well as providing just the abstracts. Claude-3-Haiku's performance on LLM-based evaluation even drops by as much as 9.3\% in overall. 
(3) \textbf{Consistent performance improvement with our framework.} Our framework shows consistent performance improvements across all three base models. Compared to retrieval-based models, our framework with Graph-Ci improves performance by 4.6\%, 2.6\%, and 3.2\% on Llama3-8B, Claude-3-Haiku, and GPT-4o, respectively. 
However, the base model's capabilities still play a dominant role. The performance of $\text{Llama3-8B}_\text{ Graph-Ci}$ is still lower than that of the abstract-based Claude-3-Haiku.

\begin{table}[t]
\setlength{\tabcolsep}{4.5pt} 

  \centering
  \footnotesize
  \begin{tabular}{l|cccc}
    \toprule
    \multirow{2}{*}{\textbf{Ours vs. Baseline}} & \multicolumn{4}{c}{\textbf{Win-Rate}} \\
    & Cov. & Log. & Rel. & Overall \\
    \midrule
    $\text{Llama3-8B}_\text{ Graph-Ci}$ vs. & \multirow{2}{*}{55\%} & \multirow{2}{*}{68\%} & \multirow{2}{*}{61\%} & \multirow{2}{*}{61.3\%} \\
    Llama3-8B + GO & \\
    $\text{Claude-3-Haiku}_\text{ Graph-Ci}$ vs. & \multirow{2}{*}{58\%}	& \multirow{2}{*}{65\%} & \multirow{2}{*}{56\%} & \multirow{2}{*}{59.7\%} \\
    Claude-3-Haiku + GO & \\
    $\text{GPT-4o}_\text{ Graph-Ci}$ vs.  & \multirow{2}{*}{56\%} &	\multirow{2}{*}{61\%} & \multirow{2}{*}{59\%} & \multirow{2}{*}{58.7\%} \\
    GPT-4o + GO & \\
    \bottomrule
  \end{tabular}
  \caption{Human evaluation results. We compare the outputs of our framework with those of the strongest baseline across three criteria (Coverage, Logic, and Relevance). Our framework consistently outperforms the strongest baseline in human evaluation.}
  \label{tab:human_eval}
  \vspace{-0.5em}
\end{table}

\subsection{Human Evaluation}
To further validate our automatic metrics, we conduct an additional human evaluation. We recruit five graduate students (master’s and PhD) specializing in artificial intelligence, all of whom have extensive experience in reviewing academic papers. We randomly select 20 papers from the artificial intelligence domain and adopt a pairwise comparison. Specifically, we pair the outputs of our framework with those of the strongest baseline, shuffle their order, and instruct human evaluators to select the better related work section based on three criteria: \textbf{Coverage}, \textbf{Logic}, and \textbf{Relevance}. We calculate the percentage of winning by our framework (Win-Rate). As shown in Table~\ref{tab:human_eval}, in human evaluation, our framework also consistently outperforms the strongest baseline (Win-Rate > 50\%) across all evaluation criteria, with the most significant improvement observed in Logic, which aligns with our LLM-based evaluation results.

\subsection{Different Strategies for Selector}
\label{sec:deciders_result}

\begin{figure*}[ht]
\centering
  \includegraphics[width=0.95\linewidth]{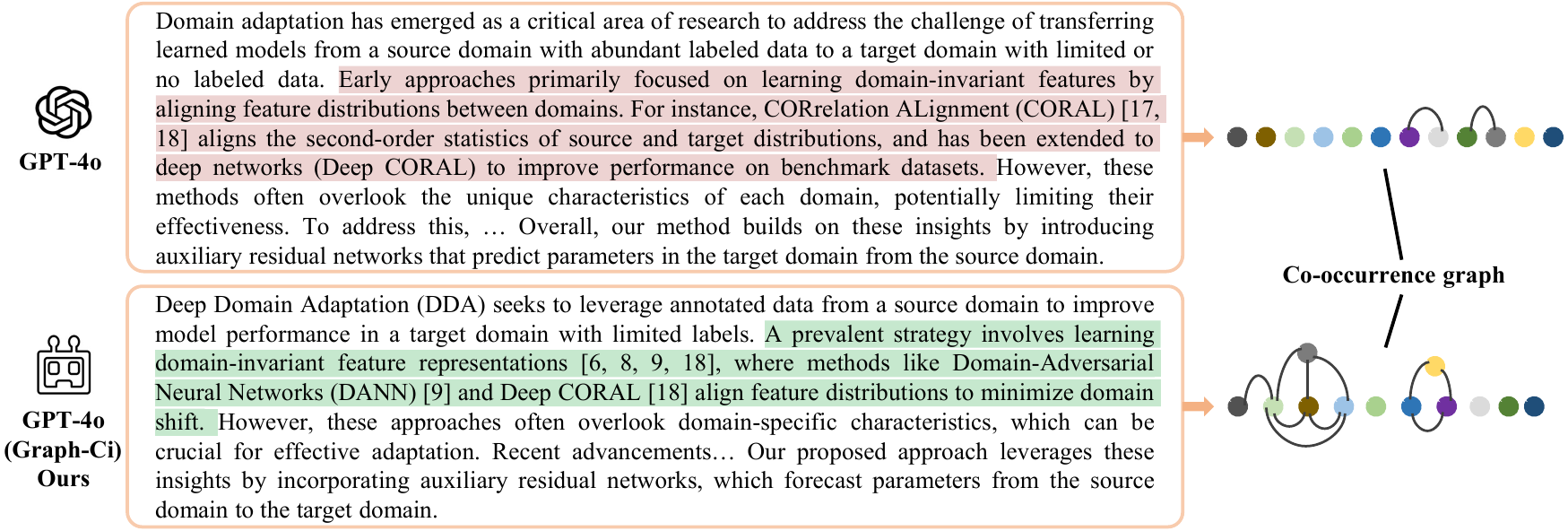}
  \caption{A case study comparing the RWS generated by GPT-4o and $\text{GPT-4o}_\text{ Graph-Ci}$. On the right, a co-occurrence graph for graph-based metrics is constructed from the generated RWS. Our $\text{GPT-4o}_\text{ Graph-Ci}$ model \protect\colorbox{mygreen}{gives a more cohesive and interrelated explanation of the references}, which is much easier for readers to follow. In contrast, GPT-4o \protect\colorbox{myred}{fails to establish connections between references}.}
  \label{fig:case_study}
\end{figure*}

We experiment with five different selector strategies across three base models. As shown in Figure~\ref{fig:deciders}, the performance trends of the five strategies are similar across the base models, following the order: SR < RR < Vanilla < Graph-Co < Graph-Ci. 
These results are expected: RR helps mitigate the potential bias in SR. Integrating LLMs into the decision-making process allows for a more intelligent selection of the reading order. Introducing the graph constraint enables the agent to more clearly capture the relationships among references. The inferior performance of Graph-Co compared to Graph-Ci may be attributed to the high connectivity of the co-occurrence graph, which imposes limited constraints on the agent's decision-making. In addition, there are minimal performance differences in Llama3-8B, which could be due to its relatively weaker capabilities, making it less sensitive to different selectors. 
The more detailed data can be found in Table~\ref{tab:deciders} in Appendix~\ref{app:detailed_res}.

\begin{table}[t]
  \centering
  \footnotesize
  \begin{tabular}{c|lc|c}
    \toprule
    \multirow{3}{*}{\textbf{Input}} & \multirow{3}{*}{\textbf{Model}}  & Avg. & Overall \\
    && Edges & {LLM} \\
    &&(Graph)&-based\\
    \midrule
     & Llama3-8B & 1.063 & 2.93 \\
     Intro.& \cellcolor{myblue}$\text{Llama3-8B}_\text{ Graph-Ci}$ & \cellcolor{myblue}1.163 & \cellcolor{myblue}3.29\\
     \multirow{2}{*}{\&}& Claude-3-Haiku & 1.452 & 3.33 \\
     &\cellcolor{myblue}$\text{Claude-3-Haiku}_\text{ Graph-Ci}$ & \cellcolor{myblue}2.413 & \cellcolor{myblue}3.41 \\
     Con.& GPT-4o & 1.033 & 3.69 \\
     & \cellcolor{myblue}$\text{GPT-4o}_\text{ Graph-Ci}$ &\cellcolor{myblue}1.735 & \cellcolor{myblue}3.71 \\
     \midrule
     \multirow{2}{*}{} & Llama3-8B & 1.088 & 3.22 \\
     &\cellcolor{myblue}$\text{Llama3-8B}_\text{ Graph-Ci}$ & \cellcolor{myblue}1.385&\cellcolor{myblue}3.31 \\
     Related& Claude-3-Haiku & 2.324 & 3.29 \\
      Work & \cellcolor{myblue}$\text{Claude-3-Haiku}_\text{ Graph-Ci}$ & \cellcolor{myblue}2.796 &\cellcolor{myblue}3.49 \\
     \multirow{2}{*}{}& GPT-4o &1.938 & 3.71\\
     &\cellcolor{myblue}$\text{GPT-4o}_\text{ Graph-Ci}$ & \cellcolor{myblue}1.918 & \cellcolor{myblue}3.73 \\
    \bottomrule
  \end{tabular}
  \caption{Performance of different models under two common input configurations. Our proposed framework consistently improves the performance of all three base models across both settings.}
  \label{tab:shorttext}
\end{table}

\subsection{Different Input Configurations}\label{sec:diff_input}
In addition to the abstracts, existing works also utilize introduction and conclusion~\cite{chen2019automatic,deng2021automatic} or RWS~\cite{xing2020automatic,ge2021baco} to represent references. We also apply our framework in these scenarios with limited portions of papers as text inputs and present the results in Table~\ref{tab:shorttext}. There are some interesting findings: (1) When limiting the task input to only the Intro. \& Con. or RWS, our framework still improves the performance of all base models. It underscores the robustness and adaptability of our framework when applied to text inputs of varying lengths. (2) A comparison between Table~\ref{tab:main_results} and Table~\ref{tab:shorttext} reveals that for both Llama3-8B and Claude-3-Haiku, providing additional sections results in a performance decline. It could stem from the relatively weaker long-text processing capabilities of these models. In contrast, for GPT-4o, the inclusion of additional sections improves its performance. (3) Even with the integration of our framework, models based on partial sections still fall short in performance compared to models utilizing the full text. It emphasizes the necessity of full-text-based RWG task. (4) Models leveraging the RWS consistently outperform those based on the Intro. \& Con. This aligns well with common academic writing practices, where the RWS of previous work is often a primary source for crafting the RWS of one’s own paper. The detailed data can be found in Table~\ref{tab:detailed_shorttext} in Appendix~\ref{app:detailed_res}.

\subsection{Case Study}
Figure~\ref{fig:case_study} presents a case study comparing the RWS generated by GPT-4o without and with our framework. The RWS generated by $\text{GPT-4o}_\text{ Graph-Ci}$ is significantly more organized, with a clearer structure and stronger connections between references. The corresponding co-occurrence graph is also notably denser. 
It gives a more cohesive and interrelated explanation of the references, which is much easier for readers to follow. In contrast, GPT-4o 
fails to establish connections between references. The explanations of individual references are overly detailed and disjointed. As a result, it is less coherent and harder for readers to grasp, which is a common struggle in previous models. By constraining the reading process to the citation graph, our $\text{GPT-4o}_\text{ Graph-Ci}$ model is better able to capture the relationships among references, resulting in a more logically structured and tightly connected output.

\section{Conclusion}
In this paper, we propose a multi-agent framework along with a graph-aware selector within the framework for full-text-based related work generation (RWG) tasks.
The framework consists of three agents: a \emph{selector}, a \emph{reader}, and a \emph{writer}, which work collaboratively to read the papers in selected order and finally generate the related work section (RWS). 
Our framework enables to optimize both the reading order and memory update. 
Our graph-aware selector can operate under the constraints of the graph to better capture the relationships among references. 
Extensive experiments demonstrate that our framework consistently improves the performance of different base models across various input configurations and the graph-aware selector based on the citation graph achieves the best performance. 
Case study reveals that our framework generates more logically coherent and tightly connected RWS. 
\section*{Acknowledgements}
This work is supported by the National Key R\&D Program of China (2023YFF0905402) and the National Natural Science Foundation of China (NSFC) (NO. 62476279, NO. U2436209). We acknowledge the anonymous reviewers for the helpful comments.

\section*{Limitations}
While our proposed framework improves graph-based metrics across different base models, indicating that the generated related work sections better capture the relationships among references, there is still a significant gap compared to human-written related work. Human-written related work can achieve an average number of edges of 9.48, whereas our best model, $\text{Claude-3-Haiku}_\text{ Graph-Ci}$, only reaches 3.24. This gap is primarily due to the model's inability to effectively handle the level of detail in different references. For example, references that could be summarized in a single sentence may be overly elaborated by the model, leading to lower coherence and relevance in the generated related work. Addressing this issue is a key focus for our future work. 

Our framework also requires that users provide a set of references in advance. Significant effort still needs to be spent on manually retrieving and selecting relevant papers. It limits the practical applicability of our method. We aim to develop a unified framework in the future where users can simply provide keywords or the citing paper, and the system will automatically retrieve the relevant papers from a vast corpus, pipelining the process of generating related work.

\section*{Ethical Statement}
Given the exponential growth of academic publications, manually curating a comprehensive and relevant related work section has become increasingly challenging and time-consuming. The RWG task aims to enhance the efficiency of scientific work by reducing the time and effort required for authors to draft the related work section of their papers. However, the misuse of automatic RWG tools could raise ethical concerns, such as the potential for the generated related work to inadvertently plagiarize content or misrepresent the details of reference papers. Therefore, the related work generated by our model is intended to serve only as a preliminary draft, helping authors save time during the writing process. Authors are still required to carefully revise and verify the output to ensure academic integrity. We believe that using such models as assistive tools rather than a replacement for thorough reading and writing can enhance the exploration of vast scientific literature. The benefits of these tools are expected to outweigh the risks, provided they are used responsibly.

\bibliography{main}
\appendix

\section{Further Experimental Analysis}
\subsection{Different Levels of Relationships}

\begin{table*}
  \centering
  \footnotesize
  \begin{tabular}{l|ccc|c}
    \toprule
    \multirow{2}{*}{Model} & Avg. & Avg. Node & Clustering & Overall \\
    & Edges & Degree & Coefficient & LLM-based\\
    \midrule
    $\text{GPT-4o}_\text{ Section}$ & 1.810 & 0.591 & 0.110 & 3.77 \\
    $\text{GPT-4o}_\text{ Graph-Ci}$ & \textbf{2.125} & \textbf{0.667} & \textbf{0.128} & \textbf{3.87} \\
    \bottomrule
 \end{tabular}
  \caption{Performance of our framework utilizing different levels of relationships.}
  \label{tab:diff_level}
\end{table*}

Our framework establishes relationships between papers. However, the selection process operates at the section level. This discrepancy may weaken the effectiveness of guiding the reading sequence. To illustrate experimentally, we construct a section-level relationship graph, denoted as \textit{Section}, where edges represent semantic similarity between different sections (i.e., their BERT embeddings exceed a similarity threshold of 0.5). We then apply the same selector mechanism to this graph while keeping all other settings unchanged. As shown in Table~\ref{tab:diff_level}, utilizing a section-level relationship graph does not guide the reading sequence of sections more effectively.

We utilize article-level relationships because these high-level relationships are crucial for composing the Related Work section, where the focus is on explaining inter-paper relationships rather than intra-paper section dependencies. Furthermore, the structure of sections within an article is generally consistent, typically following a standard organization such as Introduction, Method, and Experiments. It suggests that section-level relationships are relatively fixed. Utilizing article-level relationships proves more effective for our task.

\subsection{Different Prompt Settings}
\begin{table}
  \centering
  \footnotesize
  \begin{tabular}{l|ccc|c}
    \toprule
    \multirow{2}{*}{Model} & Avg. & Overall\\
    & Edges & LLM-based \\
    \midrule
    $\text{GPT-4o}_\text{ Graph-Ci}$ & 2.125 & 3.87 \\
    $\text{GPT-4o}_\text{ Graph-Ci}\text{-Revise}$ & 2.160 & 3.86 \\
    \bottomrule
 \end{tabular}
  \caption{Performance of our framework with different prompt settings.}
  \label{tab:diff_prompt}
\end{table}
The components of our framework is on the basis of prompts, which may vary in different prompt settings. To further evaluate the robustness of our framework, we conduct an additional experiment by significantly modifying the original prompts while preserving their core meaning. Specifically, we leverage GPT-4o to rewrite our existing prompts and then reconstruct our framework using these revised prompts, denoted as \textit{Revise}. As shown in Table~\ref{tab:diff_prompt}, the performance of our framework remains stable across different prompt settings, confirming its robustness. The prompt design may play a more critical role in single-turn interactions (e.g., Chain-of-Thought reasoning), whereas in multi-agent iterative frameworks like ours, the construction of a well-structured workflow is the key factor influencing overall effectiveness.

\subsection{Errors Analysis}
\begin{table}[t]
  \centering
  \footnotesize
  \begin{tabular}{lc}
    \toprule
    Model & Retention Ratio \\
    \midrule
    $\text{Llama3-8B}_\text{ Graph-Ci}$ & 88.39\% \\
    $\text{Claude-3-Haiku}_\text{ Graph-Ci}$ & 90.77\% \\
    $\text{GPT-4o}_\text{ Graph-Ci}$ & 96.13\% \\
    \bottomrule
 \end{tabular}
  \caption{The retention ratio of our framework (the proportion of cited papers that remain in working memory by the end of the process).}
  \label{tab:retention}
\end{table}

\begin{table}[t]
  \centering
  \footnotesize
  \begin{tabular}{lc}
    \toprule
    Model & Accuracy \\
    \midrule
    $\text{Llama3-8B}_\text{ Graph-Ci}$ & 89.86\% \\
    $\text{Claude-3-Haiku}_\text{ Graph-Ci}$ & 91.36\% \\
    $\text{GPT-4o}_\text{ Graph-Ci}$ & 94.82\% \\
    \bottomrule
 \end{tabular}
  \caption{The accuracy of information in the final working memory of our framework.}
  \label{tab:attribute_accuracy}
\end{table}
	
In our framework, the working memory is dynamically updated throughout the entire process. This iterative augmentation procedure, however, may introduce potential errors. For instance, the working memory may eventually retain only two or three highly relevant papers, or the details of Paper A may be mistakenly attributed to Paper B. Such issues—retention of cited papers and accurate attribution of information—are closely related to the instruction-following capabilities of LLMs. To mitigate these risks, we explicitly instruct the \textit{reader} to retain all cited papers and clearly indicate their corresponding IDs in the prompt. The following empirical results demonstrate that LLMs, under these guidelines, are generally able to operate with few errors.

We analyze the final working memory contents and compute the retention ratio—the proportion of cited papers that remain in working memory by the end of the process. As shown in Table~\ref{tab:retention}, in most cases, LLMs successfully retain the majority of cited papers in working memory, with stronger instruction-following capabilities leading to higher retention rates.

Similarly, we evaluate the accuracy of information in the final working memory. Using GPT-4o as a judge, we assess whether the content attributed to each cited paper in working memory is consistent with its original paper. As shown in Table~\ref{tab:attribute_accuracy}, LLMs also maintain a high level of accuracy in information attribution. Furthermore, the Relevance metric can measure whether the discussions of reference papers in the final generated related work section align with their original papers. As shown in Table~\ref{tab:main_results}, Relevance consistently scores higher than other metrics, suggesting that such attribution errors are not a dominant issue within our framework. A more critical challenge remains in accurately understanding and structuring the relationships between different papers, which is a key direction for further improvement.

\section{Section Reading Statistics}

\begingroup
\setlength{\tabcolsep}{11pt} 
\begin{table}
  \centering
  \footnotesize
  \begin{tabular}{lc}
    \toprule
    \multirow{2}{*}{Model} & Average proportion of \\
    & content read (\%) \\
    \midrule
    $\text{GPT-4o}_\text{ SR}$ & 100.00 \\
    $\text{GPT-4o}_\text{ RR}$ & 100.00\\
    $\text{GPT-4o}_\text{ Vanilla}$ & 35.27 \\
    $\text{GPT-4o}_\text{ Graph-Co}$ & 28.53 \\
    $\text{GPT-4o}_\text{ Graph-Ci}$ & \textbf{25.81} \\
    \bottomrule
  \end{tabular}
  \caption{The average proportion of content read by our framework (GPT-4o base) with different selectors.}
  \label{tab:content_proportion}
\end{table}
\endgroup

\begin{figure}
\centering
  \includegraphics[width=0.9\columnwidth]{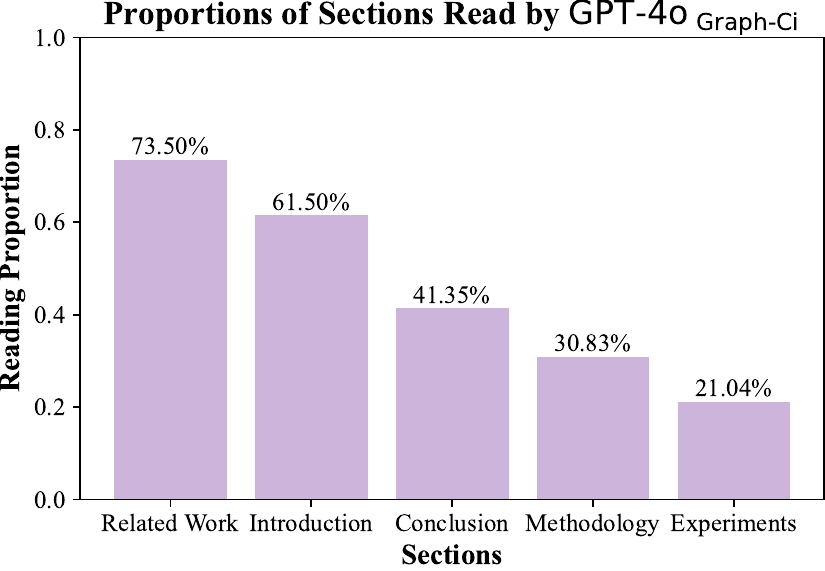}
  \caption{The proportion of sections that are selected for reading by $\text{GPT-4o}_\text{ Graph-Ci}$.}
  \label{fig:reading_statistics}
\end{figure}

We first report the average proportion of content read by our framework with different selectors (i.e., the number of sections read / total sections). As shown in Table~\ref{tab:content_proportion}, both SR and RR require reading all content. When LLMs are used for decision-making, the amount of content read is significantly reduced. 
The graph-aware selector further reduces the amount of required reading and Graph-Ci requires the least content to be read, at just 25.81\%. It further emphasizes the advantage of Graph-Ci, which achieves the highest performance while reducing both time and computational costs.

To understand what content our framework selects for reading to achieve optimal performance, we conduct an analysis of the sections read by the best-performing model, $\text{GPT-4o}_\text{ Graph-Ci}$. We categorize the sections of the papers into five categories: Introduction, Related Work, Methodology, Experiments, and Conclusion. We then calculate the proportion of these sections that are selected for reading, as shown in Figure~\ref{fig:reading_statistics}. We do not include the abstracts because we provide the abstracts of all papers for the model. The results reveal that the selector mainly selects the RWS, with a reading proportion of 73.5\%, while the Experiments sections are the least read. 
This is consistent with our experience in writing the related work section. By focusing on these high-proportion sections, researchers could reduce the reading overhead while still obtaining sufficient information to write high-quality related work.

\section{Detailed Results}
\label{app:detailed_res}
Table~\ref{tab:deciders} reports all the evaluation results for five different selector implementations across three base models (Llama3-8B, Claude-3-Haiku, and GPT-4o), representing the raw data for Figure~\ref{fig:deciders}. These additional results are consistent with the conclusions drawn in Section~\ref{sec:deciders_result}.

Table~\ref{tab:detailed_shorttext} reports all the evaluation results for the performance of different models under two common input configurations across three base models (Llama3-8B, Claude-3-Haiku, and GPT-4o). These additional results are consistent with the conclusions drawn in Section~\ref{sec:diff_input}.

\begin{table*}[h]
  \centering
  \footnotesize
  \begin{tabular}{l|ccc|cccc}
    \toprule
    \multirow{3}{*}{\textbf{Model}} & \multicolumn{3}{c}{\textbf{Graph-based Metrics}} & \multicolumn{4}{c}{\textbf{LLM-based Evaluation}} \\
    & Avg. & Avg. Node & Clustering & \multirow{2}{*}{Coverage} & \multirow{2}{*}{Logic} & \multirow{2}{*}{Relevance} & \multirow{2}{*}{Overall} \\
    & Edges & Degree & Coefficient &&&& \\
    \midrule
    $\text{Llama3-8B}_\text{ SR}$ & 0.923 & 0.413 & 0.063 & 2.70 & 3.12 & 4.02 & 3.28 \\
    $\text{Llama3-8B}_\text{ RR}$ & 0.902 & 0.433 & 0.094 & 2.70 & 3.20 & 3.96 & 3.29 \\
    $\text{Llama3-8B}_\text{ Vanilla}$ & 1.154 & 0.455 & 0.077 & 2.76 & 3.10 & 3.98 & 3.28 \\
    $\text{Llama3-8B}_\text{ Graph-Co}$ & 1.162 & 0.644 & 0.135 & 2.74 & 3.20 & 3.98 & 3.31 \\
    $\text{Llama3-8B}_\text{ Graph-Ci}$ & \textbf{1.410} & \textbf{0.651} & \textbf{0.154} & \textbf{2.80} & \textbf{3.34} & \textbf{4.18} & \textbf{3.44} \\
    \midrule
    $\text{Claude-3-Haiku}_\text{ SR}$ & 2.120 & 0.602 & 0.119 & 2.88 & 3.50 & 4.08 & 3.49 \\
    $\text{Claude-3-Haiku}_\text{ RR}$ & 2.260 & 0.617 & 0.108 & 2.92 & 3.52 & 4.12 & 3.52 \\
    $\text{Claude-3-Haiku}_\text{ Vanilla}$ & 2.720 & 0.668 & 0.117 & 2.94 & 3.50 & 4.20 & 3.55 \\
    $\text{Claude-3-Haiku}_\text{ Graph-Co}$ & 2.840 & 0.832 & 0.210 & 2.98 & 3.48 & \textbf{4.22} & 3.56 \\
    $\text{Claude-3-Haiku}_\text{ Graph-Ci}$ & \textbf{3.240} & \textbf{0.942} & \textbf{0.231} & \textbf{3.00} & \textbf{3.62} & \textbf{4.22} & \textbf{3.61} \\
    \midrule
    $\text{GPT-4o}_\text{ SR}$ & 1.760 & 0.602 & 0.106 & 3.20 & 3.78 & 4.20 & 3.73 \\
    $\text{GPT-4o}_\text{ RR}$ & 1.750 & 0.572 & 0.117 & 3.22 & 3.76 & 4.28 & 3.75 \\
    $\text{GPT-4o}_\text{ Vanilla}$ & 1.840 & 0.563 & 0.108 & 3.22 & 3.84 & 4.28 & 3.78 \\
    $\text{GPT-4o}_\text{ Graph-Co}$ & 1.900 & 0.649 & 0.123 & 3.28 & 3.74 & 4.34 & 3.79 \\
    $\text{GPT-4o}_\text{ Graph-Ci}$ & \textbf{2.125} & \textbf{0.667} & \textbf{0.128} & \textbf{3.32} & \textbf{3.86} & \textbf{4.44} & \textbf{3.87} \\
    \bottomrule
  \end{tabular}
  \caption{Performance of five different selector implementations across three base models on the OARelatedWork dataset. The best for each base model are in \textbf{bold}.}
  \label{tab:deciders}
\end{table*}

\begin{table*}[h]
  \centering
  \footnotesize
  \begin{tabular}{c|l|ccc|cccc}
    \toprule
    \multirow{3}{*}{\textbf{Input}} & \multirow{3}{*}{\textbf{Model}} & \multicolumn{3}{c}{\textbf{Graph-based Metrics}} & \multicolumn{4}{c}{\textbf{LLM-based Evaluation}} \\
    && Avg. & Avg. Node & Clustering & \multirow{2}{*}{Coverage} & \multirow{2}{*}{Logic} & \multirow{2}{*}{Relevance} & \multirow{2}{*}{Overall} \\
    && Edges & Degree & Coefficient &&&& \\    
    \midrule
     & Llama3-8B & 1.063 & 0.505 & 0.094 & 2.38 & 2.60 & 3.80 & 2.93 \\
     Intro.&\cellcolor{myblue}$\text{Llama3-8B}_\text{ Graph-Ci}$ & \cellcolor{myblue}1.163 & \cellcolor{myblue}0.522 & \cellcolor{myblue}0.124 & \cellcolor{myblue}2.66 & \cellcolor{myblue}3.12 & \cellcolor{myblue}4.08 & \cellcolor{myblue}3.29\\
     \multirow{2}{*}{\&}& Claude-3-Haiku & 1.452 & 0.525 & 0.107 & 2.52 & 3.30 & 4.18 & 3.33 \\
     &\cellcolor{myblue}$\text{Claude-3-Haiku}_\text{ Graph-Ci}$ & \cellcolor{myblue}2.413 & \cellcolor{myblue}0.776 & \cellcolor{myblue}0.164 & \cellcolor{myblue}2.76 & \cellcolor{myblue}3.40 & \cellcolor{myblue}4.08 & \cellcolor{myblue}3.41 \\
     Con.& GPT-4o & 1.033 & 0.537 & 0.117 & 3.14 & 3.68 & 4.26 & 3.69 \\
     &\cellcolor{myblue}$\text{GPT-4o}_\text{ Graph-Ci}$ & \cellcolor{myblue}1.735 & \cellcolor{myblue}0.545 & \cellcolor{myblue}0.107 & \cellcolor{myblue}3.20 & \cellcolor{myblue}3.62 & \cellcolor{myblue}4.30 & \cellcolor{myblue}3.71 \\
     \midrule
     \multirow{2}{*}{} & Llama3-8B & 1.088 & 0.442 & 0.125 & 2.52 & 3.16 & 3.98 &3.22 \\
     &\cellcolor{myblue}$\text{Llama3-8B}_\text{ Graph-Ci}$ & \cellcolor{myblue}1.385 & \cellcolor{myblue}0.534 & \cellcolor{myblue}0.115 & \cellcolor{myblue}2.76 & \cellcolor{myblue}3.14 & \cellcolor{myblue}4.04 & \cellcolor{myblue}3.31 \\
     Related& Claude-3-Haiku & 2.324 & 0.538 & 0.110 & 2.62 & 3.20 & 4.06 & 3.29 \\
      Work &\cellcolor{myblue}$\text{Claude-3-Haiku}_\text{ Graph-Ci}$ & \cellcolor{myblue}2.796 & \cellcolor{myblue}0.736 & \cellcolor{myblue}0.173 & \cellcolor{myblue}2.90 & \cellcolor{myblue}3.46 & \cellcolor{myblue}4.12 & \cellcolor{myblue}3.49\\
     \multirow{2}{*}{}& GPT-4o & 1.938 & 0.536 & 0.084 & 3.20 & 3.70 & 4.24 & 3.71\\
     &\cellcolor{myblue}$\text{GPT-4o}_\text{ Graph-Ci}$ & \cellcolor{myblue}1.918 & \cellcolor{myblue}0.560 & \cellcolor{myblue}0.117 & \cellcolor{myblue}3.20 & \cellcolor{myblue}3.68 & \cellcolor{myblue}4.32 & \cellcolor{myblue}3.73 \\
    \bottomrule
  \end{tabular}
  \caption{Performance of different models under two common input configurations. Our proposed framework consistently improves the performance of all three base models across both settings.}
  \label{tab:detailed_shorttext}
\end{table*}

\section{Prompts for Agents and Evaluation}
\label{app:prompt}

\begin{figure*}
\centering
  \includegraphics[width=\linewidth]{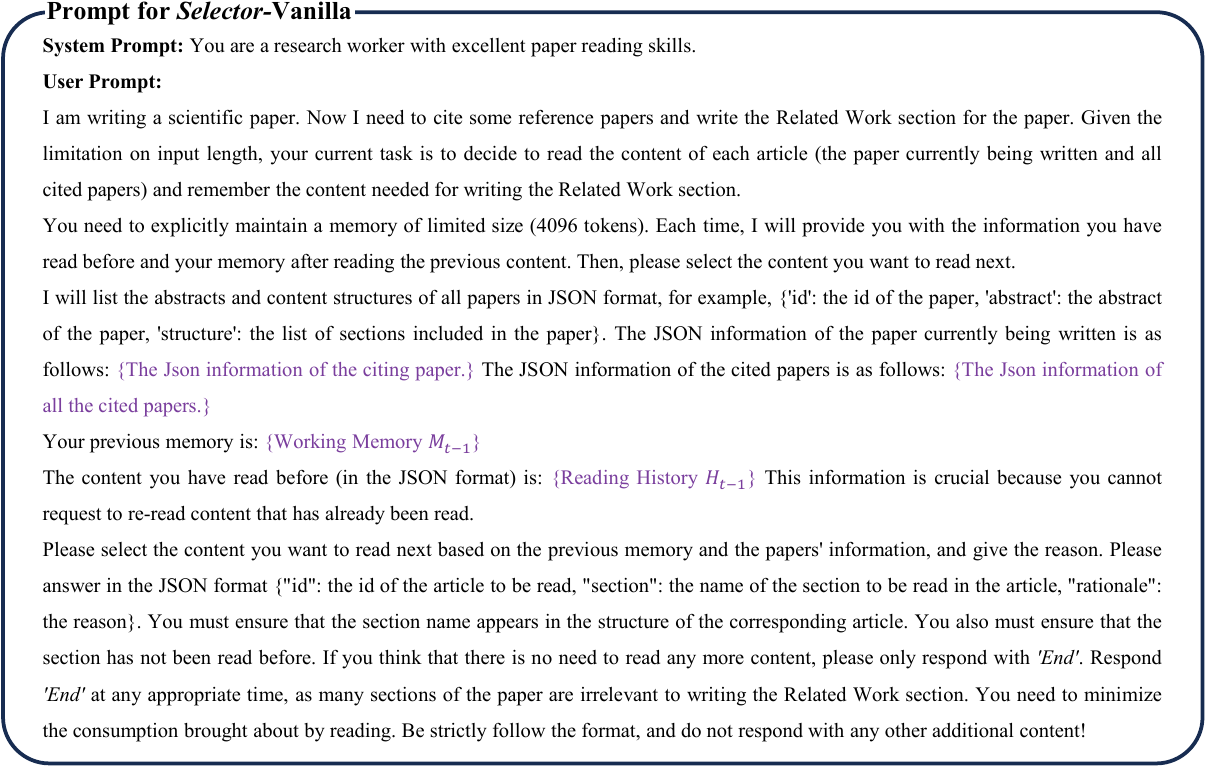}
  \caption{Prompt for \emph{Selector}-Vanilla.}
  \label{fig:prompt4vanilla}
\end{figure*}

\begin{figure*}
\centering
  \includegraphics[width=\linewidth]{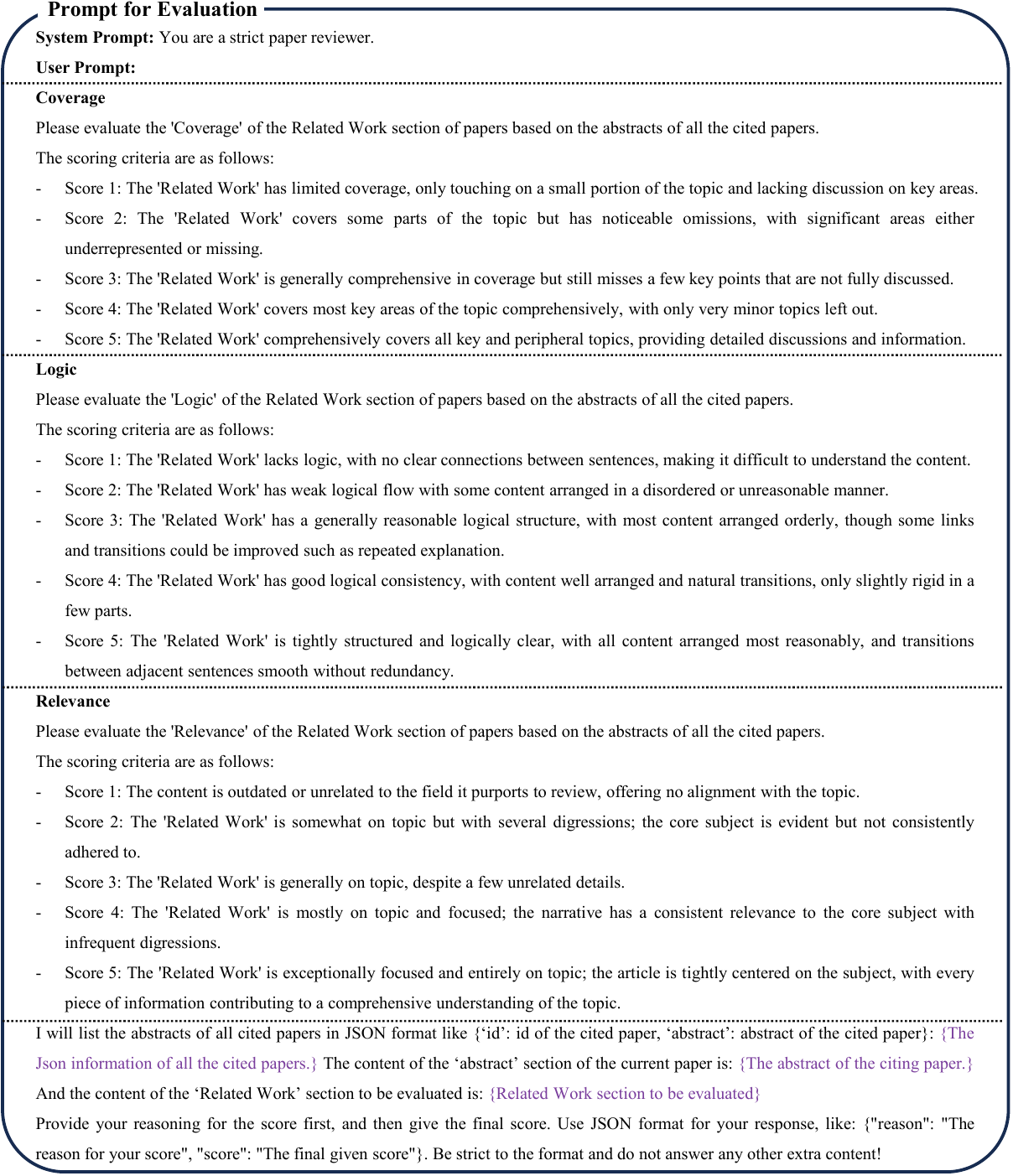}
  \caption{Prompt for Evaluation.}
  \label{fig:prompt4eval}
\end{figure*}

Figure~\ref{fig:prompt4vanilla} - Figure~\ref{fig:prompt4eval} present the detailed prompts for agents and evaluation.

\begin{figure*}[!h]
\centering
  \includegraphics[width=\linewidth]{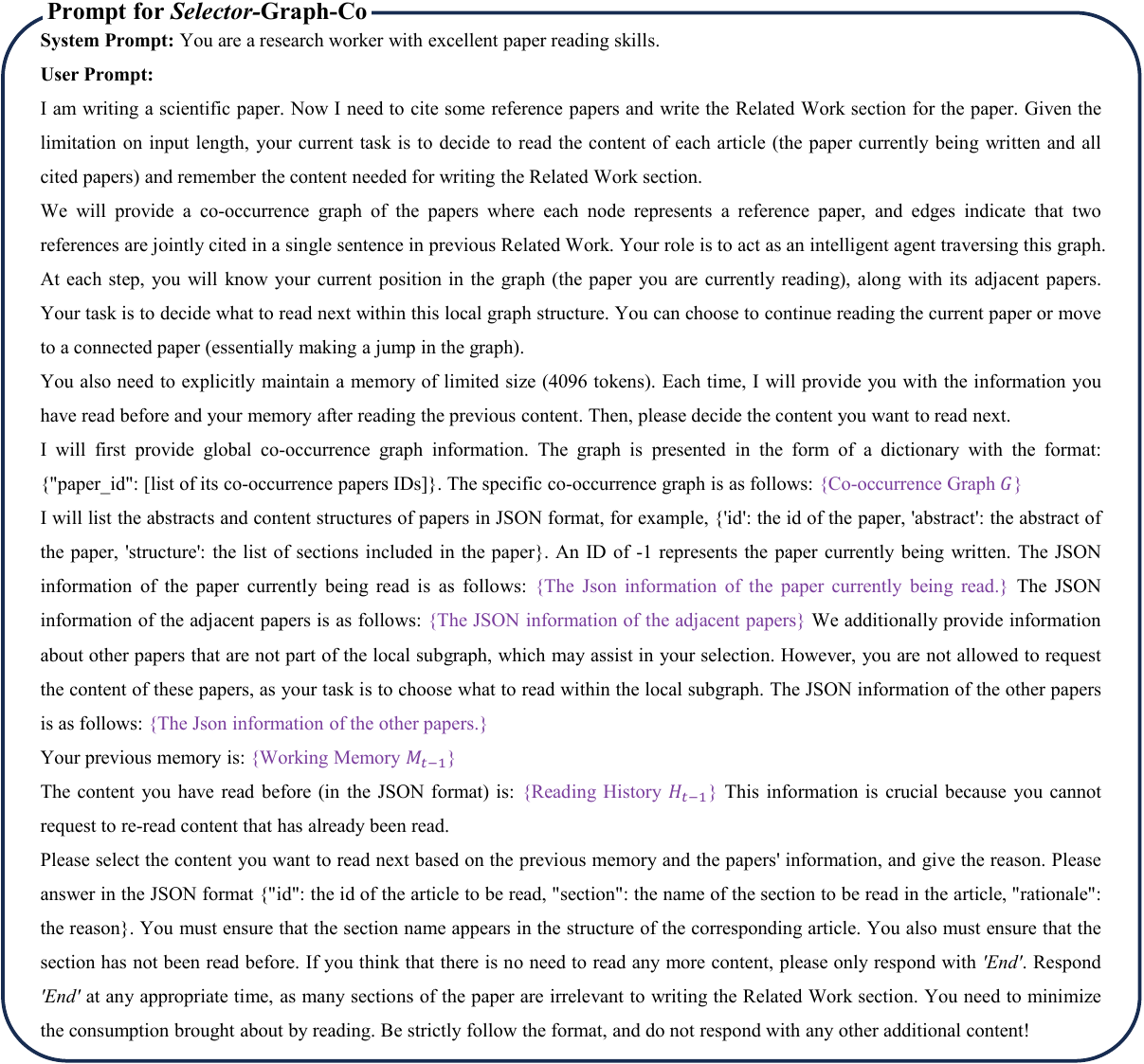}
  \caption{Prompt for \emph{Selector}-Graph-Co.}
\end{figure*}

\begin{figure*}[!h]
\centering
  \includegraphics[width=\linewidth]{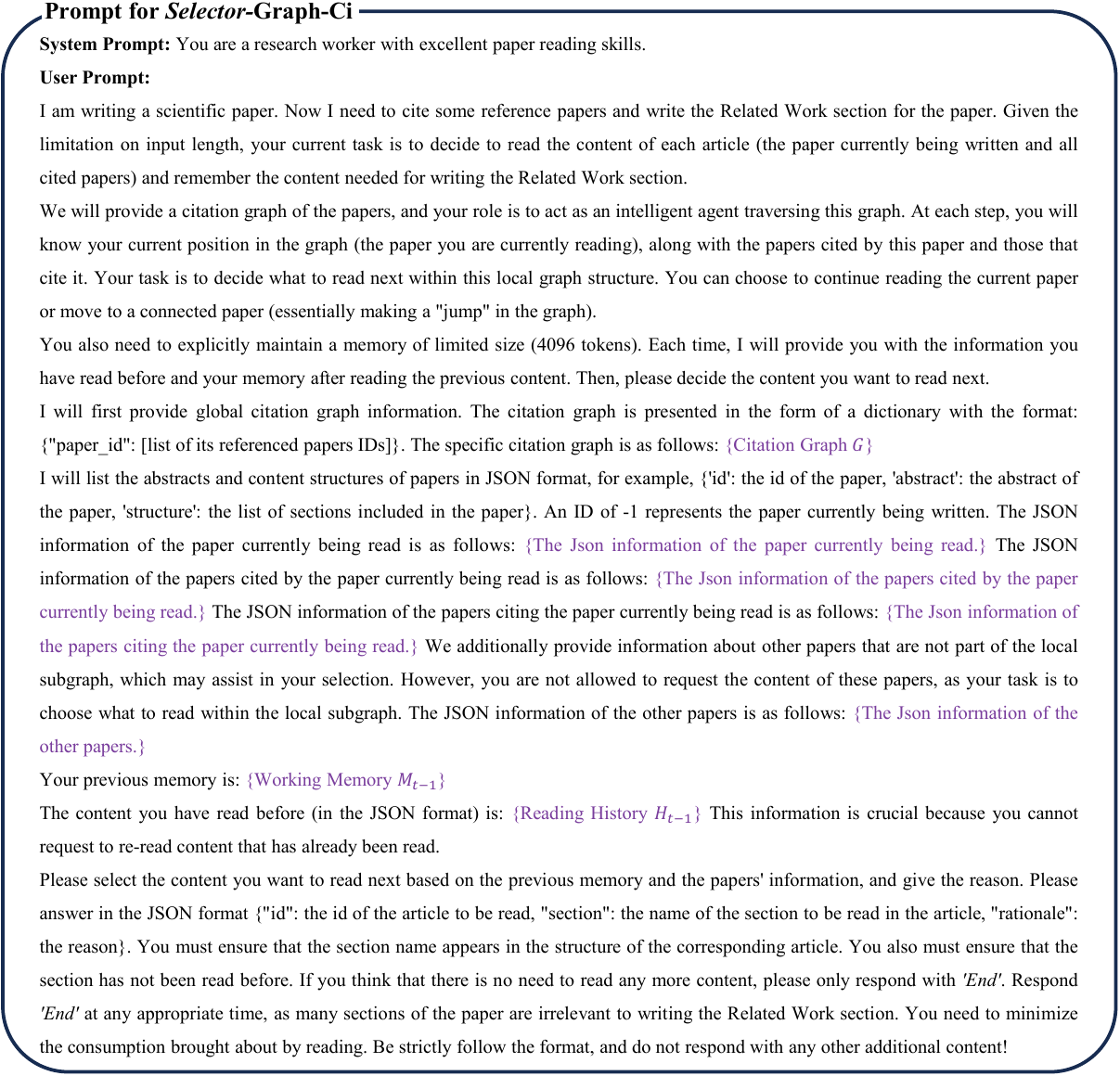}
  \caption{Prompt for \emph{Selector}-Graph-Ci.}
\end{figure*}

\begin{figure*}[!h]
\centering
  \includegraphics[width=\linewidth]{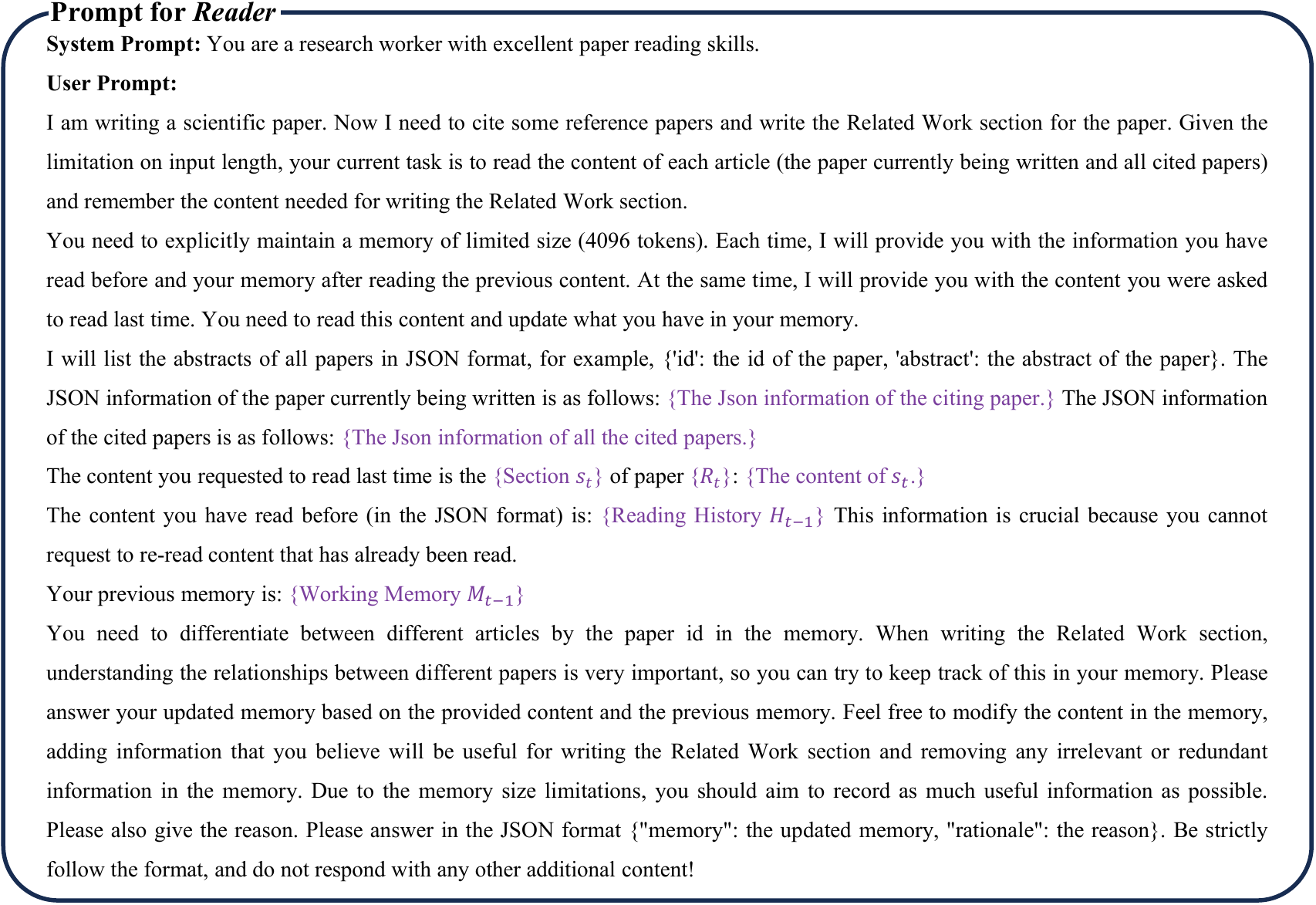}
  \caption{Prompt for \emph{Reader}.}
\end{figure*}

\begin{figure*}[!h]
\centering
  \includegraphics[width=\linewidth]{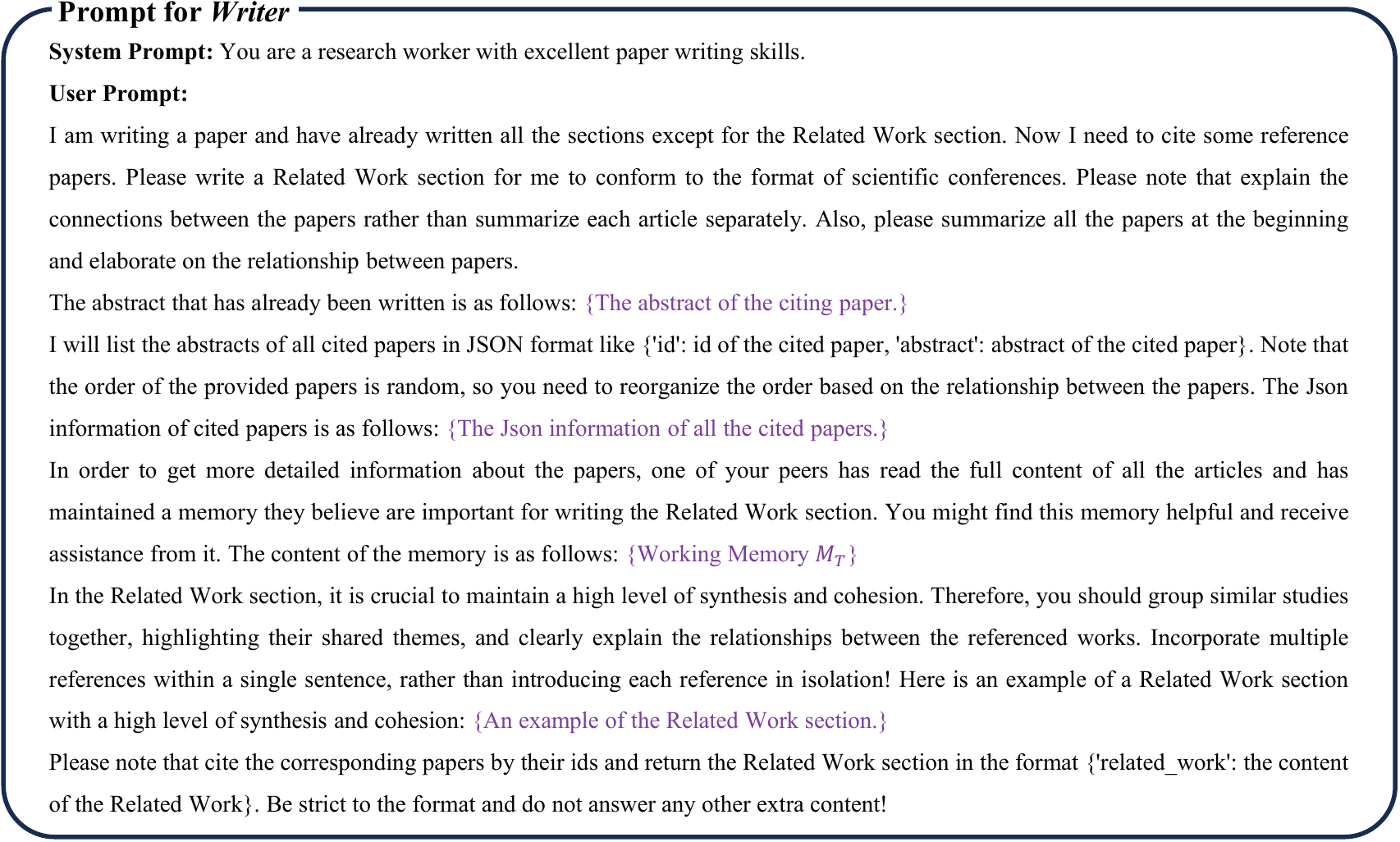}
  \caption{Prompt for \emph{Writer}.}
\end{figure*}

\end{document}